\documentclass[a4paper,fleqn]{cas-dc}

\usepackage[authoryear,longnamesfirst]{natbib}

\usepackage{tabularx,booktabs}
\usepackage{amsmath,amsfonts}
\usepackage{amssymb}
\usepackage{mathtools}
\usepackage{algorithmic}
\usepackage{algorithm}
\usepackage{array}
\usepackage[font=small,labelfont={bf}]{caption}
\usepackage{textcomp}
\usepackage{stfloats}
\usepackage{url}
\usepackage{verbatim}
\usepackage{graphicx}
\usepackage{multirow}
\usepackage{siunitx}
\usepackage{subfig}
\usepackage{lipsum}

\usepackage{natbib}
\setcitestyle{numbers,square}
\usepackage{bbm}
\usepackage{xcolor}
\usepackage{multirow}
\usepackage{makecell}
\usepackage{pifont}
\usepackage{bbding}
\usepackage{siunitx}
\usepackage{booktabs}
\usepackage{tabularx}

\definecolor{Green}{RGB}{0, 80, 0}

\definecolor{RevisedColor}{rgb}{0,0,0}

\begin{document}
\let\WriteBookmarks\relax
\def\floatpagepagefraction{1}
\def\textpagefraction{.001}

\title[mode = title]{Head Anchor Enhanced Detection and Association for Crowded Pedestrian Tracking}

\tnotemark[1]

\tnotetext[1]{
    This work was supported in part by xxx.}

\author[1,2]{Zewei Wu}[orcid=0000-0003-2289-0799]
\cortext[1]{Corresponding author:zewei.wu@mpu.edu.mo}
\cormark[1]
\author[2]{César Teixeira}[orcid=0000-0001-9396-1211]
\author[1]{Wei Ke}[orcid=0000-0003-0952-0961]
\author[3]{Zhang Xiong}[orcid=0000-0002-9421-1014]

\address[1]{Macao Polytechnic University, Macao SAR, China}
\address[2]{University of Coimbra, Coimbra 3004-531, Portugal}
\address[3]{Beihang University, Beijing 100191, China}

\begin{abstract}[S U M M A R Y]
Visual pedestrian tracking represents a promising research field, with extensive applications in intelligent surveillance, behavior analysis, and human-computer interaction. 
However, real-world applications face significant occlusion challenges. 
When multiple pedestrians interact or overlap, the loss of target features severely compromises the tracker's ability to maintain stable trajectories. 
Traditional tracking methods, which typically rely on full-body bounding box features extracted from {Re-ID} models and linear constant-velocity motion assumptions, often struggle in severe occlusion scenarios.
To address these limitations, this work proposes an enhanced tracking framework that leverages richer feature representations and a more robust motion model.
Specifically, the proposed method incorporates detection features from both the regression and classification branches of an object detector, embedding spatial and positional information directly into the feature representations.
To further mitigate occlusion challenges, a head keypoint detection model is introduced, as the head is less prone to occlusion compared to the full body.
In terms of motion modeling, we propose an iterative Kalman filtering approach designed to align with modern detector assumptions, integrating 3D priors to better complete motion trajectories in complex scenes.
By combining these advancements in appearance and motion modeling, the proposed method offers a more robust solution for multi-object tracking in crowded environments where occlusions are prevalent.
\end{abstract}


\maketitle

\section{Introduction} \label{sec:introduction}

Pedestrian tracking, as a fundamental research topic in multi-object tracking~(MOT), plays a vital role in intelligent surveillance systems and complex scene understanding. 
Recent advances in deep learning~\cite{ciaparrone2020deep,luo2021multiple} have significantly boosted the performance of {MOT} systems, enabling more accurate and robust tracking in real-world scenarios.
Modern neural network-based object detectors, such as Faster R-CNN~\cite{ren2015faster} and YOLO series~\cite{bochkovskiy2020yolov4, ge2021yolox}, have achieved remarkable breakthroughs in accuracy and recall rates. 
These detectors provide high-quality real-time observations for downstream trackers~\cite{bewley2016simple, wojke2017simple, zhang2022bytetrack, zhang2021fairmot, aharon2022bot, cao2023observation}, significantly improving overall performance of {MOT} pipelines.

\begin{figure}[htbp]
  \centering
  \includegraphics[width=0.9\linewidth]{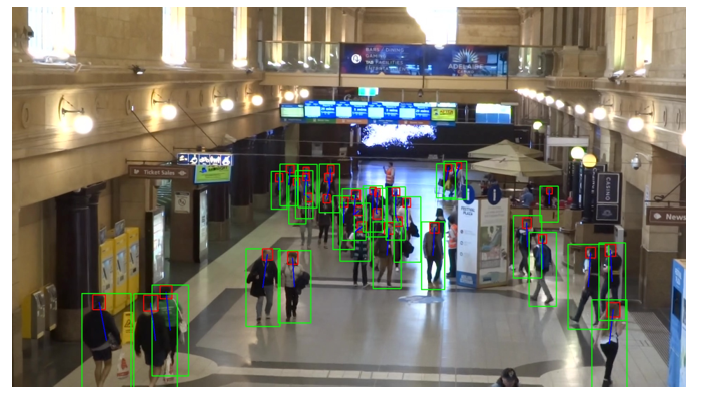}
  \captionof{figure}{%
    Illustration of occlusion events in real-world applications.
    The green bounding boxes indicate the full-body detection of pedestrians.
    The red boxes highlight the head regions of targets.
    The blue lines demonstrate the spatial relationships between head and body regions.
  }
  \label{fig:head_box}
\end{figure}

Despite these advancements,  occlusion remains one of the most persistent and challenging problems for {MOT} systems operating in crowded environments, as shown in Figure~\ref{fig:head_box}.
When pedestrians overlap or interact, complex cascaded occlusion patterns often arise, severely impairing the observability of targets and the integrity of extracted visual features.
This results in unstable tracking trajectories, higher rates of target loss, false detections, and frequent identity switches, undermining the system's reliability in real-world applications.

Traditional MOT methods typically rely on full-body bounding box detection and {Re-ID} models to extract appearance features from entire pedestrian targets.
While effective under normal conditions, this approach struggles in densely populated scenes, where full-body targets are frequently occluded, making feature extraction unreliable.
In contrast, the pedestrian head, being anatomically positioned at the highest point and less prone to occlusion, offers distinct advantages in crowded scenarios.
Head detection exhibits higher reliability in such conditions, as head regions remain visible even when the body is partially or fully occluded.
That makes them easier to re-detect and track after brief occlusions.
Therefore, head-based detection and tracking strategies have emerged as promising solutions to address occlusion challenges in high-density environments.
 
While head-based tracking methods~\cite{sundararaman2021tracking, pedersen20203d} have demonstrated advantages in occlusion-heavy scenarios, they are not without limitations.
Compared to full-body representations, head regions contain fewer discriminative features, making it difficult to distinguish between individuals in crowded scenes, particularly when multiple pedestrian heads exhibit similar appearances or sizes.
Additionally, the smaller size of head regions can reduce the robustness of feature extraction, leading to degraded tracking accuracy.
These limitations highlight the need for a more comprehensive tracking framework that integrates both head and body features.

Moreover, current {Re-ID}-based tracking methods~\cite{wojke2017simple, zhang2022bytetrack, aharon2022bot, cao2023observation} predominantly rely on full-body detection features, focusing on appearance attributes such as clothing.
However, in scenarios where pedestrians wear similar or uniform attire (e.g., schools, offices, or sports events), the discriminative capability of these methods diminishes, making it challenging to maintain identity consistency.
This underscores the limitations of conventional appearance-based representation methods in complex, real-world environments.
 
  
To address these challenges, we propose a new tracking framework that combines object detection with head keypoint regression to enhance feature representation and tracking robustness.
Inspired by anchor-free object detectors, our method leverages the geometric properties of bounding boxes alongside the spatial and positional robustness of head features to mitigate the impact of occlusion.
By integrating both full-body and head features, the proposed framework overcomes the limitations of existing approaches, ensuring precise and stable tracking even in dense crowds.
Additionally, we enhance the motion model with an iterative Kalman filtering mechanism, incorporating 3D priors to better predict and complete trajectories, especially under prolonged occlusion.
This geometry-enhanced strategy provides a more reliable foundation for maintaining target identity consistency in challenging scenarios, while preserving computational efficiency.
By addressing the limitations of both traditional full-body tracking and head-based approaches, our method delivers a robust solution for multi-object tracking in complex, crowded environments with severe occlusion.


\section{Related work} \label{sec:related_work}

\subsection{Person Detection}

Person detection has seen significant advancements over the past decades, transitioning from traditional handcrafted feature-based methods to modern deep learning approaches.
Early methods, such as Histogram of Oriented Gradients ({HOG}), demonstrated reasonable performance in constrained scenarios but struggled with occlusions, complex poses, and varying lighting conditions.
The advent of deep learning revolutionized person detection, with detectors like Faster R-CNN \cite{ren2015faster} and YOLO series \cite{redmon2016you, glenn_jocher_2021_5563715,bochkovskiy2020yolov4, ge2021yolox} achieving unprecedented detection accuracy alongside real-time performance.
These methods rely on highly expressive {CNN} to extract robust features, enabling effective pedestrian localization even in challenging environments.

Modern person detection methods can be broadly categorized into full-body detectors and part-based detectors. 
Full-body detectors like CenterNet~\cite{duan2019centernet} and YOLOX \cite{ge2021yolox}, excel in standard scenarios but suffer in crowded scenes due to frequent occlusions, where overlapping bodies compromise feature extraction.
Part-based methods, particularly those focusing on head detection \cite{sundararaman2021tracking,idrees2015detecting,shao2018crowdhuman}, provide better robustness to occlusions since the head is often less obstructed than the body.
However, head-based methods face challenges in identity discrimination, as the head region contains fewer distinct features compared to full-body representations.


To address these limitations, we propose a new approach that combines head keypoint regression with full-body detection, integrating the complementary strengths of both paradigms.
Our method builds upon recent advances in anchor-free object detection while introducing innovative strategies for handling severe occlusions and maintaining identity consistency in high-density environments.

\subsection{Pedestrian Tracing}

Pedestrian tracking has also seen significant evolution, progressing from traditional motion-based methods to modern hybrid approaches that integrate motion and appearance cues.
Traditional tracking methods primarily relied on handcrafted motion models such as Kalman filtering~\cite{Kalman} and particle filtering.
While effective in controlled environments, these methods struggled with challenges such as occlusions, abrupt motion changes, and the lack of robust appearance features in complex scenes.

With the rise of deep learning, modern tracking methods can be categorized into two primary streams: 
Appearance-based methods, which leverage deep neural networks to extract discriminative visual features for target representation.
Notable examples include DeepSORT~\cite{wojke2017simple} and FairMOT~\cite{zhang2021fairmot}, which combine {Re-ID} features with tracking-by-detection frameworks to achieve state-of-the-art performance.
Hybrid methods, which integrate appearance features with motion cues for enhanced robustness.
For instance, methods like~\cite{wang2022extendable} and~\cite{zhang2020long} utilize spatial-temporal modeling to predict motion trajectories while maintaining appearance consistency, particularly in highly occluded scenarios.

The tracking-by-detection paradigm has become the dominant framework in modern {MOT} systems.
In this paradigm, object detection and data association are treated as separate modules, allowing for flexible integration of state-of-the-art detectors and association strategies.
While this modular approach offers advantages in terms of computational efficiency and real-time performance, it faces several key challenges:
i) Detection quality directly influences tracking performance, with occlusion and overlapping targets degrading feature reliability.
ii) Feature representation is often insufficiently robust under severe occlusions, leading to degraded target identity consistency.
iii) Identity switches frequently occur in crowded scenes, particularly when targets share similar appearances or clothing.

To overcome these challenges, our work proposes a novel tracking framework that enhances target representation by combining head and full-body features.
We integrate head keypoint regression into the tracking pipeline to improve robustness against occlusion, while maintaining the modularity and efficiency of the tracking-by-detection paradigm.
Furthermore, we incorporate an iterative Kalman filtering mechanism enhanced with 3D priors to improve motion trajectory prediction under prolonged occlusions.
By addressing the limitations of current approaches, our method delivers more accurate and stable tracking in crowded environments with severe occlusion and appearance ambiguity.

\subsection{Head Enhanced Target Representation}

Despite the effectiveness of detector-based {Re-ID} features, target representation under occlusion remains a critical challenge that demands further optimization.
Occlusion events hurt target appearance and motion state observability, significantly degrading the reliability of target representation in tracking systems.
Figure~\ref{fig:head_box} demonstrates the occlusion challenges in pedestrian tracking scenarios. In crowded scenes, while full-body detections often suffer from severe occlusions, head regions maintain higher visibility. 
This observation suggests that head region features could provide complementary information for robust target representation under occlusion conditions.
 
Compared to full-body bounding boxes, head bounding boxes have a lower probability of being occluded. 
Moreover, since the head is a rigid body with minimal deformation, its features demonstrate stronger robustness. 
Those characteristics have been applied in early crowd detection~\cite{peng2018detecting, shao2018crowdhuman} and related literature~\cite{sundararaman2021tracking, pedersen20203d}, followed by researchers proposing tracking methods based on head detection. 
While these methods partially alleviate occlusion issues, their performance remains limited due to the small feature regions of heads and the limited discriminative power of head geometric shapes.

Based on these observations, we propose that utilizing head-enhanced features based on body bounding boxes can further optimize target representation capabilities.
Through this approach, we can increase the importance of head region features within the full-body representation while maintaining sufficient discriminative power even under occlusion conditions.
Notably, the body bounding boxes still effectively provide information about target body shape and spatial relationships.
 
\begin{figure} [htbp]
	\centering
	\includegraphics[width=0.99\linewidth]{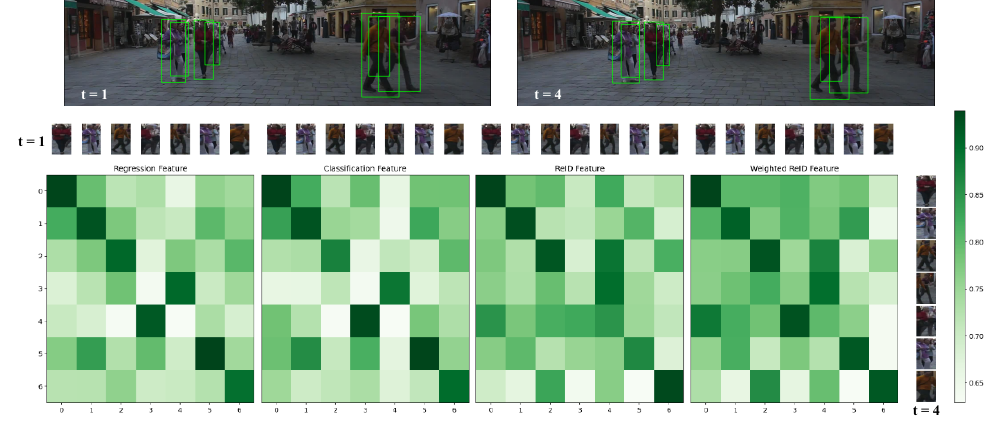}
	\caption{
        Illustration of similarity matrices for different feature types.
        Here we demonstrate four different feature types: (a) classification features, (b) regression features, (c) {Re-ID} features, and (d) an upper-body weighted {Re-ID} feature representation.
    }
	\label{fig:reid_cls_reg}
\end{figure}

To validate our hypothesis, we conducted verification experiments by extracting regression, classification and {Re-ID} features from YOLOX detector and PCB {Re-ID} model on pedestrian detections. The experiments were performed on adjacent frames from MOT17-02~\cite{milan2016mot16} dataset. 
We further incorporated a control group with weighted {Re-ID} features emphasizing the upper one-third region containing head information. 
Figure~\ref{fig:reid_cls_reg} presents the experimental results.

For clarity in our discussion, we refer to these four subfigures from left to right as $a-d$.
The Fig.~\ref{fig:reid_cls_reg}-$a$ and $b$ represent the cross-frame target similarity matrices derived from detector-based features. 
Similarity matrix in Fig.~\ref{fig:reid_cls_reg}-$a$ demonstrate remarkably consistent results, achieving peak similarity scores for correct target matches.
The matrix in Fig.~\ref{fig:reid_cls_reg}-$b$ captures the influence of spatial relationships on feature similarity and exhibits high sensitivity to spatial distance. 
Specifically, even when two patches are visually similar, their similarity score remains low if they are spatially distant. Conversely, patches that are spatially close may receive high similarity scores, even if their visual features differ significantly.
This indicates the strong discriminative power of detector-based features in target association.

When comparing Fig.~\ref{fig:reid_cls_reg}-$c$ and $d$, we focus particularly on the patch pair $(4,3)$ which presents a typical occlusion scenario that the pedestrian's body is partially occluded while the head region remains visible. In Fig.~\ref{fig:reid_cls_reg}-$c$, which utilizes conventional {Re-ID} features, we observe that the similarity score for patch pair $(4,0)$ incorrectly exceeds that of $(4,3)$. This error clearly demonstrates how {Re-ID} features can be adversely affected by foreground occlusions.

The classification and regression features shown in Fig.~\ref{fig:reid_cls_reg}-$a$ and $b$ exhibit latent superior robustness, maintaining reliable similarity scores despite the occlusion. 
This observation strongly supports our hypothesis regarding the effectiveness of detector-based features for robust target representation. The results suggest that these features can potentially offer more stable performance in challenging occlusion scenarios compared to traditional {Re-ID} approaches.

\begin{figure} [htbp]
	\centering
	\includegraphics[width=0.99\linewidth]{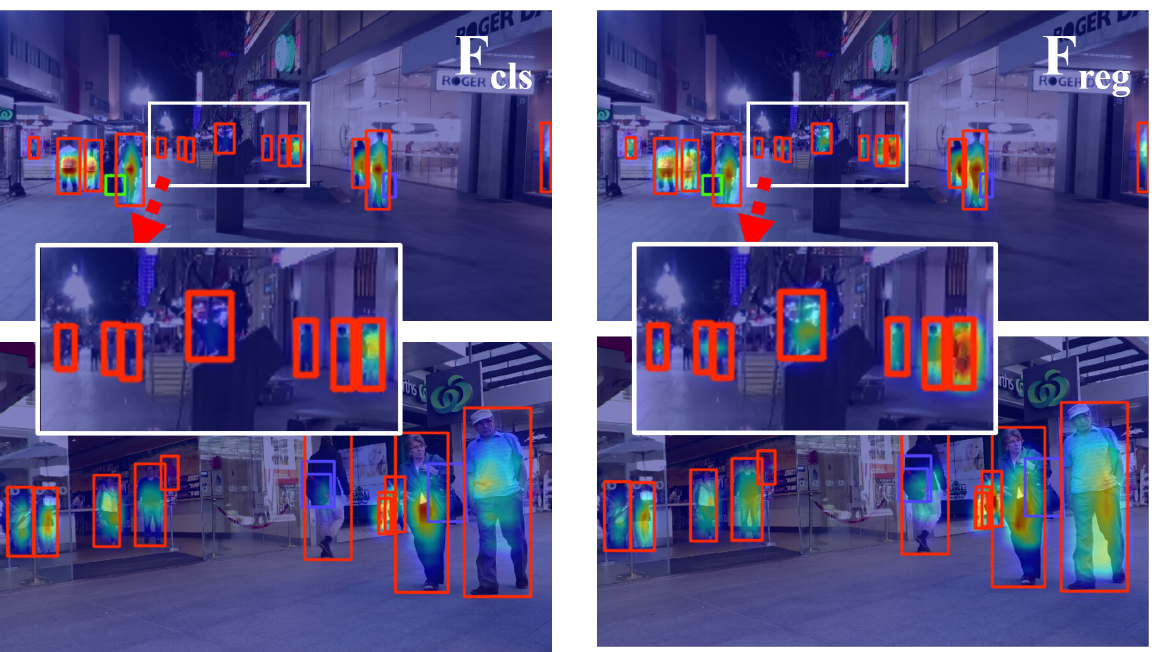}
	\caption{
        Illustration of Heatmaps from different features using LayerCAM.
    }
	\label{fig:cam_cls_reg}
\end{figure}

\section{FocusTrack}

In this section, we introduce FocusTrack, a robust {MOT} framework designed to address occlusion challenges in crowded pedestrian environments. 
The framework consists of two main components: i) an enhanced detection model incorporating head keypoints and full-body bounding boxes, and ii) a tracking module that leverages these features for robust identity association and motion prediction.
Our key innovations include the integration of head keypoint detection into an existing anchor-free architecture, a new feature selection strategy combining head and full-body cues, and a motion model incorporating iterative Kalman filtering with 3D priors interpolation.

\subsection{Proposed Detection Model}

\begin{figure} [htbp]
	\centering
	\includegraphics[width=0.9\linewidth]{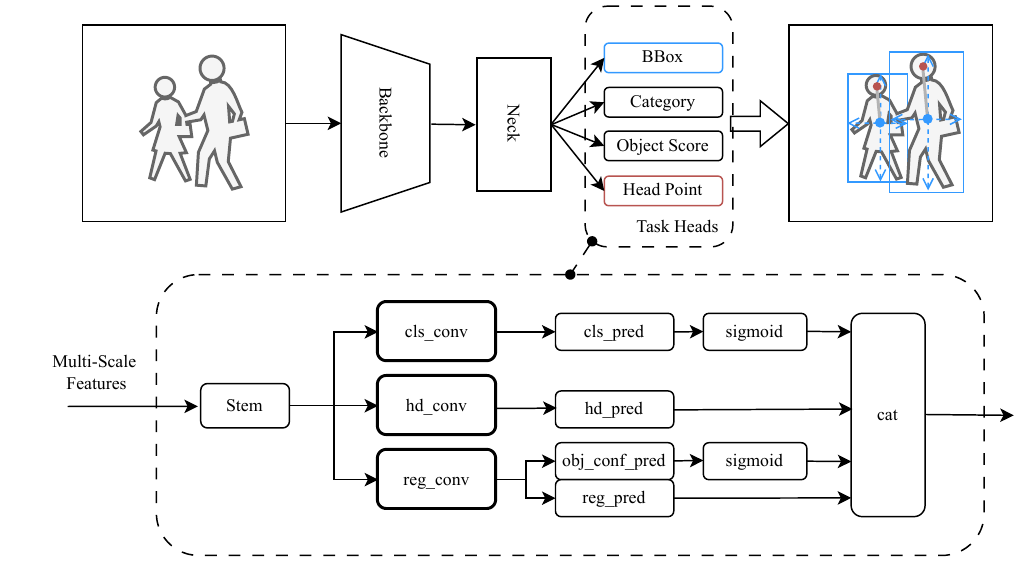}
	\caption{Illustration of the detector model.
    The backbone network employs CSPDarknet53 architecture, while the neck network utilizes PAFPNet structure. As illustrated in the figure, the detection heads are decoupled into separate branches for classification, regression, and head keypoint prediction.
    The head keypoint is defined as $[x,y,v]$, where $(x,y)$ represents the spatial coordinates of the head position and $v$ denotes the visibility score.
    }
	\label{fig:head_detector}
\end{figure}
 
\paragraph{Model Architecture}
The detection model forms the visual front end of FocusTrack, enabling accurate localization of both full-body bounding boxes and head keypoint.
Our detection model builds upon the YOLOX architecture, a state-of-the-art anchor-free object detector. The model is enhanced by adding a head keypoint prediction branch to the detection head, enabling simultaneous prediction of full-body bounding boxes and head center coordinate. 
The overall architecture is shown in Figure~\ref{fig:head_detector}.
The backbone network uses CSPDarkNet53, while the neck employs a PAFPNet structure for multi-scale feature fusion. 
The detection head is decoupled into multiple branches for classification, bounding box regression, and head keypoint prediction.

For an input image $I \in \mathbb{R}^{H \times W \times 3}$, the feature extractor model outputs predictions at three different scales $s \in \{8, 16, 32\}$. 
At each scale, the feature map $F_s \in \mathbb{R}^{H/s \times W/s \times C}$ is processed by the detection head to generate predictions $\hat{Y}_s$:
\begin{equation}
    \hat{Y}_s = \{\hat{y}_i\}_{i=1}^{N_s}, \quad \hat{y}_i = (c_i, b_i, p_i)
\end{equation}
where $N_s$ is the number of predictions at scale $s$, $c_i \in \mathbb{R}^{1}$ represents the classification logits (only 1 category), $b_i = (x, y, w, h)$ denotes the bounding box parameters, and $p_i = (x_\text{head}, y_\text{head}, v_\text{head})$ represents the head keypoint coordinates and visibility score.
Specifically, the head regression branch is modified to output both head keypoint and visibility predictions through separate {CNN} layers.
Here, head keypoint regression is a center localization of object head. 

For the head keypoint detection branch, we adopt a decoupled design approach rather than predicting head offsets from bounding box centers. While predicting head positions relative to box centers could establish explicit relationships between body and head detections, this approach has notable limitations. Head occlusions or truncations would make such relative predictions unreliable. Moreover, bounding box centers do not consistently align with body centers, and the spatial relationship between box and head centers follows no fixed pattern. Therefore, an independent detection branch provides more robust and flexible head keypoint localization.
 
To effectively incorporate head position constraints into the detector's label assignment process, we propose a novel assignment strategy that considers both bounding box and head keypoint information.

\paragraph{Label Assignment Strategy}
In our case, the label assignment strategy extends the YOLOX approach to incorporate head keypoint detection, ensuring effective multi-task learning for both object detection and head localization. 
The process begins with foreground candidate selection, where a foreground mask is generated to identify anchors whose center points fall inside the ground truth bounding boxes and their central regions.

The overall matching cost $\mathcal{C}_{\mathrm{assign}}$ for each anchor-GT pair classification cost, combines IoU cost, and a positional constraint:
\begin{equation}
	\mathcal{C}_{\mathrm{assign}} = \mathcal{C}_{\mathrm{cls}} + \alpha \cdot \mathcal{C}_{\mathrm{IoU}} + \beta \cdot \mathcal{C}_{\mathrm{dist}}
\end{equation}
where $\alpha$ and $\beta$ are weighting factors.
To be specific, the classification cost $ \mathcal{C}_{\mathrm{cls}}$ is computed by applying binary cross-entropy between the predicted class probabilities and the one-hot encoded ground truth labels, providing a measure of classification accuracy during the matching process.
This is achieved by evaluating both geometric and positional constraints, ensuring only relevant anchors are considered. 
For each valid anchor, the pairwise {IoU} cost $\mathcal{C}_{\mathrm{IoU}}$ between the predicted bounding boxes and the ground truth boxes is computed, denoted as:
\begin{equation}
\mathcal{C}_{\text{IoU}}(b_\text{pred}, b_\text{gt}) = -\log(\text{IoU}(b_\text{pred}, b_\text{gt}) + \epsilon),
\end{equation}
where $b_\text{pred}$ and $b_\text{gt}$ represent the predicted and ground truth bounding boxes, respectively; 
$\epsilon$ is a small constant to prevent numerical instability.
The positional constraint $\mathcal{C}_{\mathrm{dist}}$ is an indicator function assigning a large penalty if an anchor is outside the center region. 
By setting $\beta$ to a high value (e.g., $10^5$), such anchors are effectively excluded from matching.

Similar to YOLOX, the dynamic K-matching strategy is employed to ensure each ground truth box is assigned a sufficient number of positive anchors. 
This strategy ensures a balanced assignment of positive samples for both object-level and keypoint-level supervision, enabling the model to learn accurate bounding box regression, keypoint localization, and class prediction. 

\paragraph{Loss Function}
The loss function in our study is designed to jointly optimize object detection and keypoint localization. 
It combines multiple components, including bounding box regression, objectness prediction, classification, and keypoint regression, to achieve effective multi-task learning. 
The total loss is defined as:
\begin{equation}
	\mathcal{L}_{\text{total}} = \mathcal{L}_{\text{cls}} + \mathcal{L}_{\text{box}} + \mathcal{L}_{\text{head}} 
\end{equation}
where each term corresponds to a specific task, involving classification, bounding box regression, and head keypoint regression.

For the classification task, BCE is also used to compare the predicted class probabilities with the one-hot encoded ground truth classes:
\begin{equation}
\mathcal{L}_{\text{cls}} = \frac{1}{N_{\text{fg}}} \sum_{i \in \text{fg}} \sum_{c=1}^{C} \text{BCE}(\hat{p}_{i,c}, p_{i,c}),
\end{equation}
where $N_{\text{fg}}$ is the number of foreground samples, $C$ is the number of classes, $\hat{p}_{i,c}$ is the predicted probability for class $c$, and $p_{i,c}$ is the ground truth class label.
In this work, we only consider a single pedestrian category.
For the bounding box regression task, we consider a {IoU} loss and objectness loss to evaluate the alignment between predicted bounding boxes and ground truth boxes, and the confidence of whether an anchor corresponds to an object, respectively.
It is expressed as:
\begin{equation}
	\mathcal{L}_{\text{box}} = \underbrace{\frac{1}{N_{\text{fg}}} \sum_{i \in \text{fg}} \mathcal{L}_{\text{IoU}}(\hat{b}_i, b_i)}_{\text{bounding box loss}} + \underbrace{\frac{1}{N_{\text{fg}}} \sum_{i \in \text{anchors}} \text{BCE}(\hat{o}_i, o_i)}_{\text{objectness loss}},
\end{equation}
where $\mathcal{L}_{\text{IoU}}$ represents the CIoU~\cite{zheng2021ciou} loss. 
This term penalizes misalignment between predicted and ground truth bounding boxes.
$\hat{b}_i$ and $b_i$ are the predicted and ground truth bounding boxes.
$\hat{o}_i$ is the predicted objectness score, and $o_i$ is the ground truth objectness label.
Moreover, an optional L1 loss is applied to further refine the bounding box predictions:
\begin{equation}
\mathcal{L}_{\text{L1}} = \frac{1}{N_{\text{fg}}} \sum_{i \in \text{fg}} \|\hat{b}_i - b_i\|_1,
\end{equation}
Keypoint regression is implemented using a custom loss function that evaluates both the precision of keypoint localization and the visibility of keypoint. 
The loss $\mathcal{L}_{\text{head}}$ is defined as:
\begin{equation}
\mathcal{L}_{\text{head}} =  \underbrace{\|{k_\text{pred}} - k_\text{gt}\|_2^2}_\text{head keypoint loss} + \underbrace{\text{BCE}(v_{\text{pred}}, v_{\text{gt}})}_{\text{visibility loss}},
\end{equation}
where $\hat{k}_i$ and $k_i$ are the predicted and ground truth keypoints, respectively. 
Similar to bounding boxes, an optional L1 loss is applied for keypoint refinement.

The final loss is a weighted sum of all components. 
This design ensures balanced optimization for object detection and keypoint localization, enabling the model to achieve high accuracy across both tasks.

\subsection{Tracker Design}

Feature representation serves as a fundamental component in modern {DOT} frameworks. While conventional approaches typically employ separate networks for detection and {Re-ID} feature extraction, this dual-network architecture introduces computational redundancy and increased system complexity. We observe that the rich hierarchical features learned by modern anchor-free detectors like YOLOX can be effectively leveraged for both detection and tracking tasks.

Building upon this observation, we propose a varient SORT framework that utilizes detection features, including head keypoint information, to replace traditional {Re-ID} models. 
Furthermore, we enhance the framework with improved motion modeling and trajectory completion mechanisms to address the challenges posed by crowded scenarios and severe occlusions.
The following paragraphs detail our complete tracking pipeline, shown as Algorithm~\ref{algo:iterkfkp}, encompassing cost computation methodology, association strategy, iterative motion prediction, and trajectory completion through dimensional lifting techniques.

\begin{algorithm}[thbp] 
    \caption{Procedure of FocusTrack}\label{algo:iterkfkp}
    \small
    \textbf{Input}:
    Video frames $\left\{I_{t}\right\}_{t=1}^{T}$,
    a patience window $w$,
    a gating threshold $g$,
    a iteration error threshold $\eta$,
    a set of matched detections $\mathcal{M}$,
    a set of unmatched detections $\mathcal{U}$.
    \begin{algorithmic}[1]
        \STATE Initialize $\mathcal{M} = \emptyset$, $\mathcal{U} = \mathcal{D}$.
        \FOR{timestep $t \leftarrow 1: T$} %
            \STATE Perform detection on image frame $I_t$ to obtain detections $\left\{d_{i}\right\}_{i=1}^{N}$ and head points;
            \STATE Obtain features $F_t^\text{det}$ from detector features through coordinate query;
            \STATE Construct cost matrix ${C}_{\text{assoc}}$;
            \STATE Solve minimum cost matching using Hungarian algorithm with cost ${C}_{\text{assoc}}$;

            \STATE $ \mathcal{M} \leftarrow \mathcal{M} \cup\left\{(\tau_{i}, d_j) \mid {C}_{\text{assoc}}[i, j] \leq g\right\} $
            \STATE $ \mathcal{U} \leftarrow \mathcal{U} \backslash\left\{{d_j}\mid \operatorname{min} \{{C}_{\text{assoc}}[i, j]\}_{i=1}^N > g\right\} $
            
            \FOR {$ d_{i} \text{ in } \mathcal{U}$}
                \IF {$d_{i}$ is a newborn track}
                    \STATE Create a new track $\tau$, $\tau.count \leftarrow 0$;
                    \STATE Initialize a new kalman filter;
                \ELSIF {$\tau.count ++ \geqslant w$}
                    \STATE Eliminate $\tau$;
                \ENDIF
            \ENDFOR
            \FOR {$\tau \text{ in } \mathcal{M}$}
            \STATE $\mathcal{T} \leftarrow \mathcal{T} \bigcup \{\tau\}$
            \ENDFOR
            \WHILE{measurement residuals $\mathbf{r}$ > $\eta$}
                \STATE update parameters of kalman filters;
            \ENDWHILE
        \ENDFOR
    \end{algorithmic}
\end{algorithm}

\paragraph{Feature Selection}
As discussed earlier, occlusion presents a significant challenge in crowded scenes, particularly for pedestrian tracking where appearance features are highly susceptible to occlusion effects. 
This challenge is not merely a limitation of appearance representation models.
In occluded detection boxes, the foreground target often dominates the region, yet the appearance model struggles to faithfully reconstruct the target's features.

To address this limitation, we propose a novel feature selection approach that leverages deep learning detector features. This approach is motivated by the observation that deep detection models, trained on whole images, inherently learn position, shape and appearance characteristics while optimizing their primary detection objectives. We hypothesize that such models can provide valuable target representations, particularly geometric and positional features that remain effective under occlusion conditions. By incorporating head detection, we utilize these detector features as a new feature selection mechanism.

The feature extraction process is straightforward and efficient. 
Leveraging the anchor-free detection architecture, we employ inverse coordinate mapping ${M}$ to locate corresponding feature maps: 
\begin{equation}
    \begin{aligned}
    {M}: &\mathbb{R}^2 \rightarrow \mathbb{R}^2, (x_c, y_c) \mapsto  (i,j), \\&i = \left\lfloor\frac{x_c}{s}\right\rfloor,
    j = \left\lfloor\frac{y_c}{s}\right\rfloor,
    \end{aligned}
    \end{equation}
where $s\in \{8,16,32\}$ represents the stride of feature maps corresponding to three scales $P3$, $P4$, and $P5$. 
Further, we also obtain features from the nearest classification and regression prediction layers corresponding to the box center coordinates.
The mapping function ${M}$ projects the target center coordinates from the input image to integer index positions on the feature maps. Through floor operation, the feature map indices are guaranteed to be integers, ensuring proper alignment between image and feature space.

For full-body bounding boxes, features are extracted from the nearest classification and regression prediction layers corresponding to the box center coordinates. 
Similarly, for head keypoints, features are selected from the closest head keypoint prediction layer.

\paragraph{Cost Matrices Construction}
Following the SORT-like tracking methods, we formulate the target association as a minimum-cost bipartite matching optimization problem:
\begin{equation}
    \begin{aligned}
    \arg &\min_{x} \sum_{i,j} c_{ij}x_{ij}, \\ 
    \text{s.t.} &\sum_i x_{ij} \leq 1, \quad \sum_j x_{ij} \leq 1,
    \end{aligned}
\end{equation}
where \( x_{ij} \) denotes the binary assignment variable and \( c_{ij} \) represents the association cost between track \( i \) and detection \( j \).

The tracking performance primarily depends on the association cost matrix \( C_\text{assoc} \), which measures the similarity between existing tracks and new detections. In our FocusTrack framework, we design a comprehensive cost matrix that integrates both appearance and motion cues:
\begin{equation}
    C_\text{assoc} = w_\text{app} \cdot C_{\text{app}} + w_\text{mot} \cdot C_{\text{mot}}
\end{equation}
where $C_{\text{app}}$ represents the appearance cost derived from detector features, $C_{\text{mot}}$ denotes the motion-based cost, and $w$ is a balancing coefficient that weights these complementary terms. The appearance cost function integrates multiple feature components: classification logits and regression features from the detector's output branches, combined with head keypoint embeddings through cosine similarity measurement. 
For motion modeling, the cost is calculated as the Euclidean distance between the current detection and the predicted state estimated by a Kalman filter, providing temporal consistency in the tracking process.

\paragraph{Iterative Motion Prediction}
In this study, the target state is modeled as an 8-dimensional vector 
\begin{align}
    \mathbf{x} = [u,v,a,h,\delta{u},\delta{v},\delta{a},\delta{h}]^T
    \end{align}
where $(u,v)$ represents the target's center position, $a$ denotes the aspect ratio, $h$ indicates the height, and their corresponding velocities are denoted by $(\delta{u},\delta{v},\delta{a},\delta{h})$. 
The state transition matrix $F$ models the constant velocity motion, while the measurement matrix $H$ maps the state vector to observations. The measurement noise covariance $R$ and process noise covariance $Q$ are determined empirically based on detection uncertainty and motion dynamics respectively. The standard Kalman filter performs state prediction through
\begin{align}
    \hat{\mathbf{x}}_{t|t-1} &= {F}_t\hat{\mathbf{x}}_{t-1|t-1} \\
    {P}_{t|t-1} &= {F}_t{P}_{t-1|t-1}{F}_t^T + Q_t 
    \end{align}
and update through:
\begin{align}
{K}_t & ={P}_{t \mid t-1} {H}_t^{\top}\left({H}_t {P}_{t \mid t-1} {H}_t^{\top}+{R}_t\right)^{-1} \\
\hat{\mathbf{x}}_{t \mid t} & =\hat{\mathbf{x}}_{t \mid t-1}+{K}_t\left(\mathbf{z}_t-{H}_t \hat{\mathbf{x}}_{t \mid t-1}\right) \label{eq:state_update}\\
{P}_{t \mid t} & =\left({I}-{K}_t {H}_t\right) {P}_{t \mid t-1} \label{eq:covariance_update}
\end{align}

However, this linear constant velocity model can accumulate significant errors when targets undergo occlusion or non-linear motion. 
The measurement update step becomes critical, especially since modern detectors provide highly accurate position measurements with relatively small uncertainties. 

Modern object detectors, when properly trained, can provide highly accurate position measurements with relatively low measurement uncertainty.
To better leverage these precise measurements while handling non-linear motion, we propose an iterative Kalman update procedure.
Starting with $\delta\mathbf{x}_k^{(0)} = 0$, for each iteration $j$, we compute:
\begin{align}
K_t^{(j)} &= P_{t|t-1}H_t^T\left(H_tP_{t|t-1}H_t^T + R_t\right)^{-1} \\
\delta\mathbf{x}_t^{(j+1)} &= \delta\mathbf{x}_t^{(j)} + K_t^{(j)} (\underbrace{\mathbf{z}_t - h\left(\hat{\mathbf{x}}_{t|t-1} \oplus \delta\mathbf{x}_t^{(j)}\right) + H_t\delta\mathbf{x}_t^{(j)}}_{j\text{-th observation residuals }r_t^{(j)}})
\end{align}

After convergence at iterations, the final state $\hat{\mathbf{x}}_{t|t}$ and covariance $P_{t|t}$ updates with Equation~\ref{eq:state_update} and Equation~\ref{eq:covariance_update}. 
The number of iterations is determined adaptively based on the convergence criterion:
\begin{equation}
\|\delta\mathbf{x}_t^{(j+1)} - \delta\mathbf{x}_t^{(j)}\| < \epsilon \cdot \|\mathbf{r}_t^{(0)}\|
\end{equation}
where $\epsilon$ is a small threshold (typically 0.01) and $\|\mathbf{r}_t^{(0)}\|$ is the initial residual magnitude. This ensures efficient convergence while maintaining tracking accuracy.


\begin{figure} [htbp]
	\centering
	\includegraphics[width=8.0 cm]{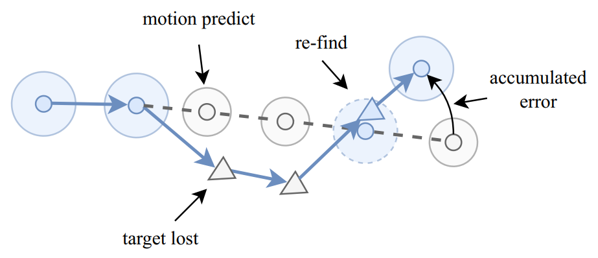}
	\caption{Illustration of the false due to strong motion prior.
    }
	\label{fig:false}
\end{figure}

\paragraph{Trajectory Completion via Dimensional Lifting}
For non-real-time scenarios, we propose a trajectory completion mechanism based on dimensional lifting to handle occlusions and missing detections. 
The approach lifts 2D coordinates $(x,y)$ into 3D space $(x,y,z)$ through pseudo-depth estimation:
\begin{equation}
    z = f_\text{depth}(x, y, \theta)
\end{equation}
where $f_\text{depth}(\cdot)$ denotes the depth estimation function and $\theta$ represents camera parameters. 
While many depth estimation methods require substantial computational resources and extensive annotated data or precise camera calibration parameters, we adopt a simplified yet effective approach. By assuming targets move on a common ground plane with the camera mounted at an acute angle, a typical setup for surveillance scenarios that depth can be approximated using following equation:
\begin{equation}
    z= d + \frac{1}{y + \eta}
\end{equation}
where $d$ represents a minimum depth estimation and $y$ denotes the vertical coordinate of the bounding box's bottom edge.
$\eta$ is a constant to adjust division by zero.
This simplified approach is based on an intuitive depth assumption: in many cases, the perceived depth of an object in an image tends to exhibit an inverse relationship with the object's projected location, particularly along the vertical axis of the image plane.
This straightforward estimation provides sufficient depth cues for trajectory completion while maintaining computational efficiency.

The 3D trajectory interpolation is then performed using Lie group modeling:
\begin{equation}
    T(t) = T_1 \exp(\omega \log(T_1^{-1}T_2))
\end{equation}
where $T_1, T_2$ are trajectory keyframes, $\omega \in [0,1]$ is the interpolation parameter, and $\exp(\cdot), \log(\cdot)$ are the exponential and logarithm maps on the SE(3) Lie group.
This dimensional lifting approach enables more accurate trajectory interpolation by considering the geometric constraints of human motion in 3D space.

\section{A New Pedestrian Tracking Dataset}

To evaluate our proposed method, we constructed a comprehensive pedestrian tracking dataset (called FT25) combining real-world and synthetic data sources. The dataset integrates carefully re-annotated sequences from the MOT20 benchmark dataset~\cite{dendorfer2020mot20}, which features complex occlusion patterns and varying crowd densities. For these sequences, we performed meticulous annotation refinement by adding head keypoint labels and optimizing bounding box annotations. Additionally, we generated synthetic data using a customized simulation pipeline based on the CARLA platform, enabling precise control over environmental conditions and automatic ground truth generation. This dual-source approach leverages both the complexity of real-world scenarios and the controllability of synthetic environments, facilitating thorough evaluation of our tracking framework under diverse conditions.

\subsection{Real-world Data Collection}
\begin{figure} [htbp]
	\centering
	\includegraphics[width=8.0 cm]{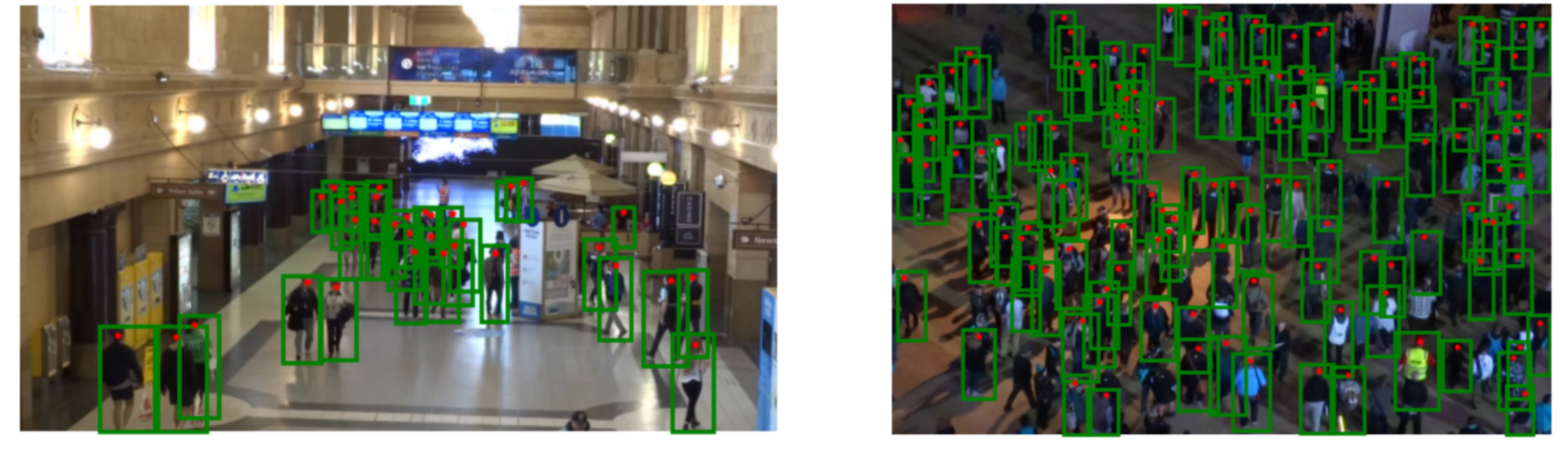}
	\caption{Illustration of the annotated examples from FT25-01 and FT25-02 sequences.
    }
	\label{fig:FT25_egs}
\end{figure}

The first part of dataset consists of two sequences: FT25-01 and FT25-02, shown as Figure~\ref{fig:FT25_egs}, containing $429$ and $3,315$ images respectively, for a total of $3,744$ images. 
Each image is annotated with head keypoint information and target bounding boxes to describe the position and structure of objects within the images. 
Based on the annotation distribution, the dataset is further divided into training and validation sets, comprising $1,873$ and $1,871$ images respectively. 
The corresponding number of samples (instances with keypoint annotations) are $211,580$ and $175,770$ for the training and validation sets. 
The complete training set, without partition, contains all $3,744$ images with a total of $387,350$ annotated samples.

\subsection{CARLA-based Data Collection Pipeline}

\begin{figure} [htbp]
	\centering
	\includegraphics[width=0.99\linewidth]{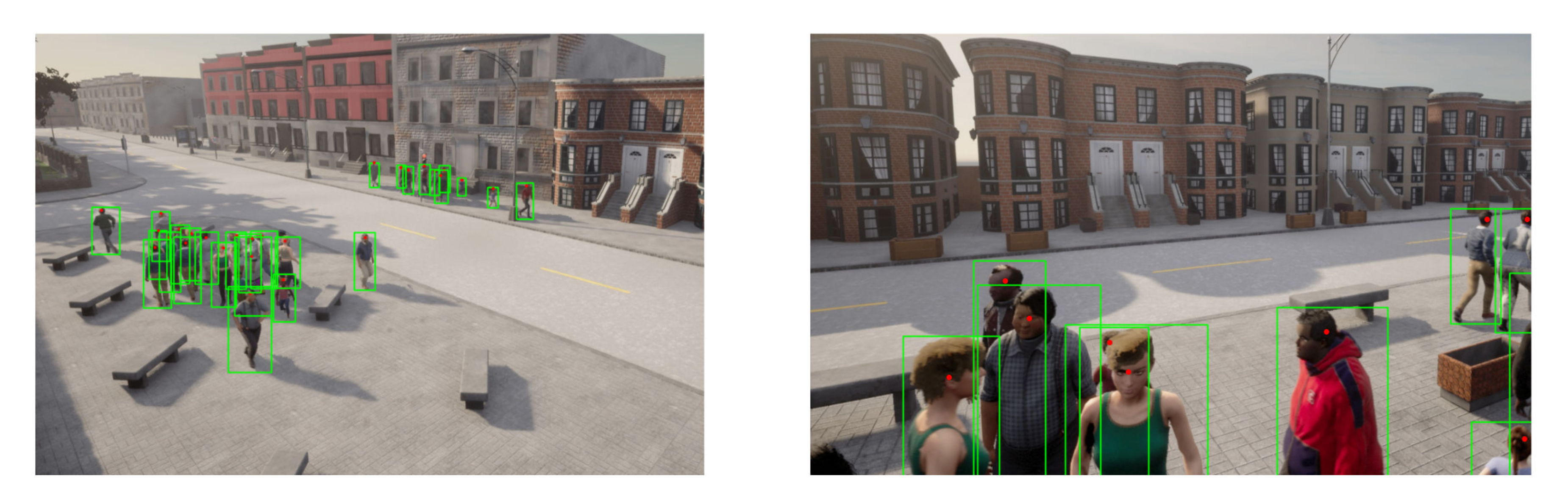}
	\caption{Illustration of the annotated examples from FT25-03 and FT25-04 sequences.
    }
	\label{fig:data_ft25_e1}
\end{figure}

To comprehensively evaluate our proposed method, we constructed a simple simulated dataset FT25-03, FT25-04 using the CARLA simulation platform, illustrated in Figure~\ref{fig:data_ft25_e1}. 
The training and validation sets contain 498 and 350 images respectively, with comprehensive annotations totaling 16,434 and 10,819 instances.
CARLA provides a efficient urban environment simulation with precise control over scene parameters and automatic ground truth generation capabilities, making it an ideal platform for developing and validating pedestrian tracking algorithms.
The simulation environment was configured using CARLA version 0.9.13. 
We carefully selected multiple urban maps with diverse architectural styles, ranging from dense city centers to open suburban areas, to ensure sufficient variety in the collected data. 
The environmental conditions were systematically varied to create a comprehensive dataset, including different weather conditions (clear, cloudy, and rain), times of day, and multiple camera viewpoints. 
Scene complexity was controlled by varying pedestrian densities between 50 to 150 agents per scene.

\begin{figure} [htbp]
	\centering
	\includegraphics[width=0.8\linewidth]{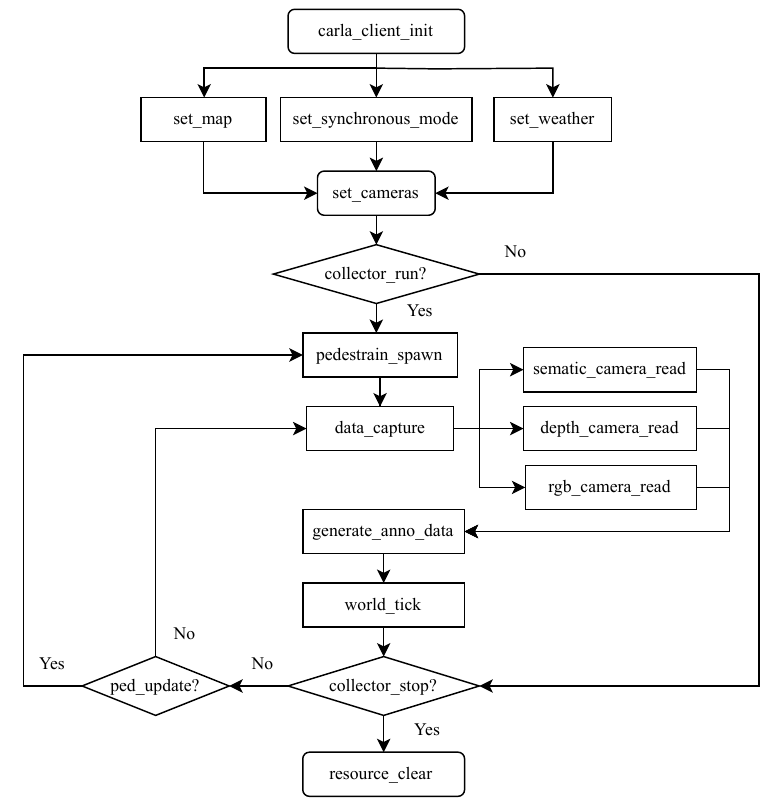}
	\caption{Illustration of the systematic pipeline of our dataset collection.
    This process consists of initialization, pedestrian generation, data acquisition, and subsequent processing stages.
    }
	\label{fig:data_pipeline}
\end{figure}

As demonstrated in Figure~\ref{fig:data_pipeline}, our data collection pipeline implements a systematic approach to generate high-quality training data through automated Python scripts. 
The pipeline operates on a client-server architecture, where the CARLA simulation server first initializes the environment, followed by client connections and synchronized data collection scripts. 
Three observation cameras were strategically mounted at preset positions in the scene, configured to simultaneously capture RGB images, depth maps, and instance segmentation masks at 20 frames per second, ensuring comprehensive multi-modal data collection.
The process begins with scene configuration through automated scripts that spawn pedestrians with randomized appearances and trajectories. 
These scripts ensure natural pedestrian behaviors while maintaining control over crowd density and movement patterns. 

\begin{figure}[htbp]
	\centering
	\includegraphics[width=8.0 cm]{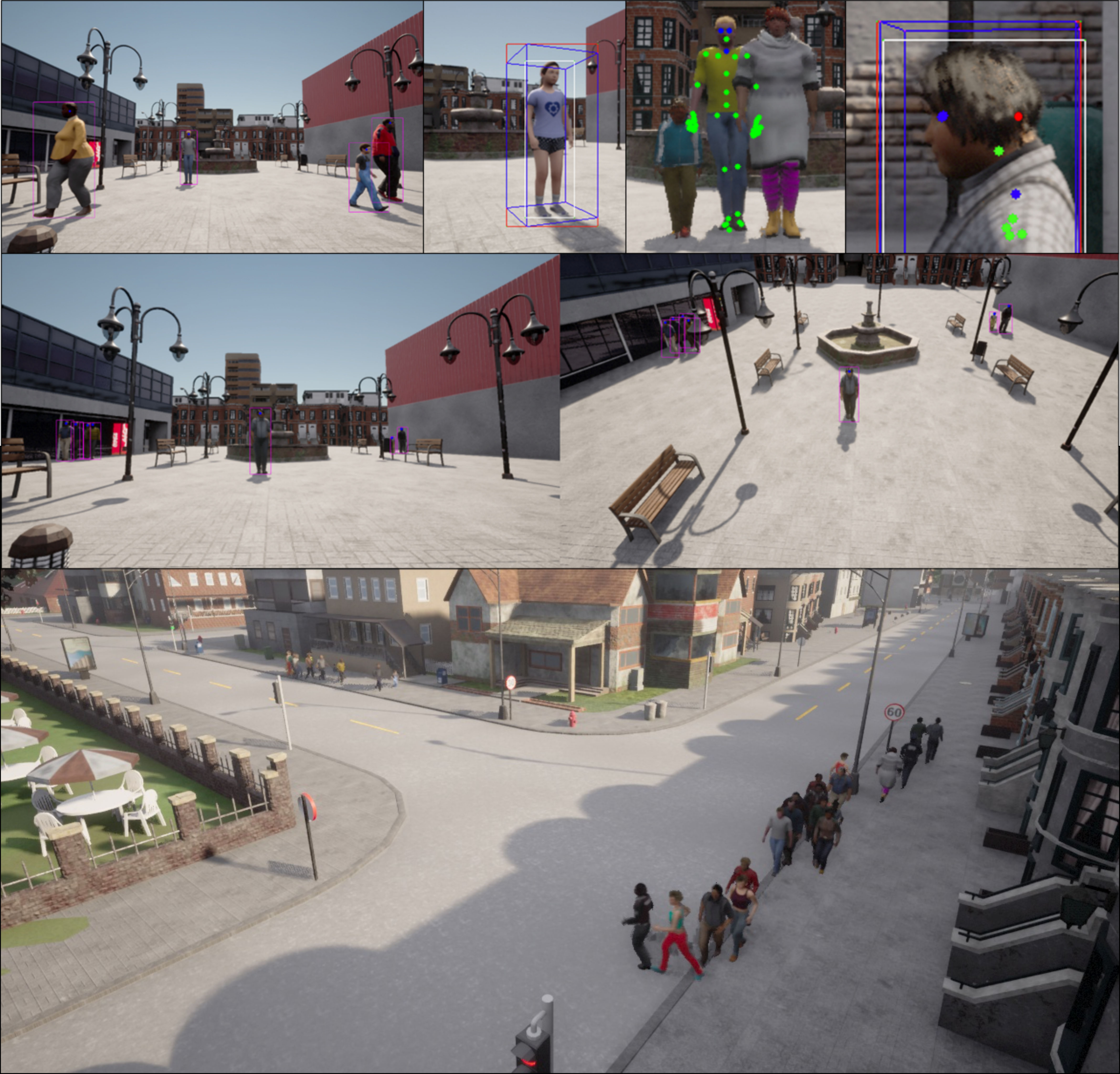}
	\caption{Illustration of examples from our simulated scenario.
    The figure illustrates diverse viewpoints and scene samples from our dataset, along with corresponding annotation information provided for comprehensive evaluation.
    The data collection pipeline also supports the generation of diverse task-specific datasets, with flexible customization of scene configurations, weather conditions, annotation formats, and other parameters.
    }
	\label{fig:data_examples}
\end{figure}

Examples from our simulated dataset are illustrated in Figure~\ref{fig:data_examples}.

\begin{table}[htbp]
    \centering
    \caption{Main detector training parameters of detector in use.
    }
    \label{tab:params_ch3}
    \renewcommand\arraystretch{1.3}
    \setlength\tabcolsep{2.5pt}
    \begin{tabular}{c|c}
    \hline
        Parameter & Value \\ \hline
        Initial weight
        & from ByteTrack~\cite{zhang2022bytetrack}  \\ \hline
        Input size
         & $800 \times 800$  \\ \hline
        Text size
         & $800 \times 800$  \\ \hline
        Random size
         & (18, 32)  \\ \hline
        Batch size
         &  16 \\ \hline
        Optimizer
         &  SGD \\ \hline
        Max epoch
        &  300 \\ \hline
        w/o augmentation 
        &  15 epoch\\ \hline
        Augmentation strategy
        &  Mosaic, Mixup, HSV, Flip  \\ \hline
        NMS theshold
        &  0.45  \\ \hline
        Confidence theshold
        &  0.25  \\ \hline
        Visibility theshold
        &  0.5 \\ \hline
        Software platform
        &  Pytorch 1.9~\cite{paszke2019pytorch}, Ubuntu 22.04 OS  \\ \hline
        CPU
        &  Intel 4214R CPU  \\ \hline
        GPU
        &  NVIDIA RTX A6000   \\ \hline

    \end{tabular}
\end{table}

\section{Experiment}
In this section, we conduct extensive experiments to evaluate the effectiveness of our proposed FocusTrack. 
 
\subsection{Experimental Setup}

We evaluate our proposed method on three datasets: MOT17, MOT20, and our newly introduced FT25 dataset. 
MOT17~\cite{milan2016mot16} and MOT20~\cite{dendorfer2020mot20} are widely-used benchmark datasets from MOTChallenge, containing diverse pedestrian tracking scenarios with complex occlusion patterns. 
These datasets serve as standard evaluation platforms for assessing tracking performance under real-world conditions.

For performance evaluation, we adopt standard multi-object tracking metrics including CLEAR MOT~\cite{bernardin2008evaluating} (like MOTA $\uparrow$, IDs $\downarrow$, FP$\downarrow$, FN$\downarrow$) and IDF1 $\uparrow$, which provide comprehensive measures of tracking accuracy, precision, and identity preservation capabilities.

The visual frontend of our tracking system builds upon the YOLOX anchor-free architecture, enhanced with head keypoint detection capabilities as previously described. 
To ensure fair comparison, our detector is only fine-tuned on the FT25 training set with frozen 2D detection layers, maintaining consistent 2D bounding box detection performance across all comparative trackers. Throughout our experimental evaluation, we compare our approach against both classical and state-of-the-art detection methods. To rigorously assess the contribution of our proposed enhancements, we conduct experiments both with and without the head keypoint detection component.
Detailed training parameters are presented in Table~\ref{tab:params_ch3}.

\subsection{Impact of Detector Feature Respresentation}

\begin{figure} [htbp]
	\centering
	\includegraphics[width=8.0 cm]{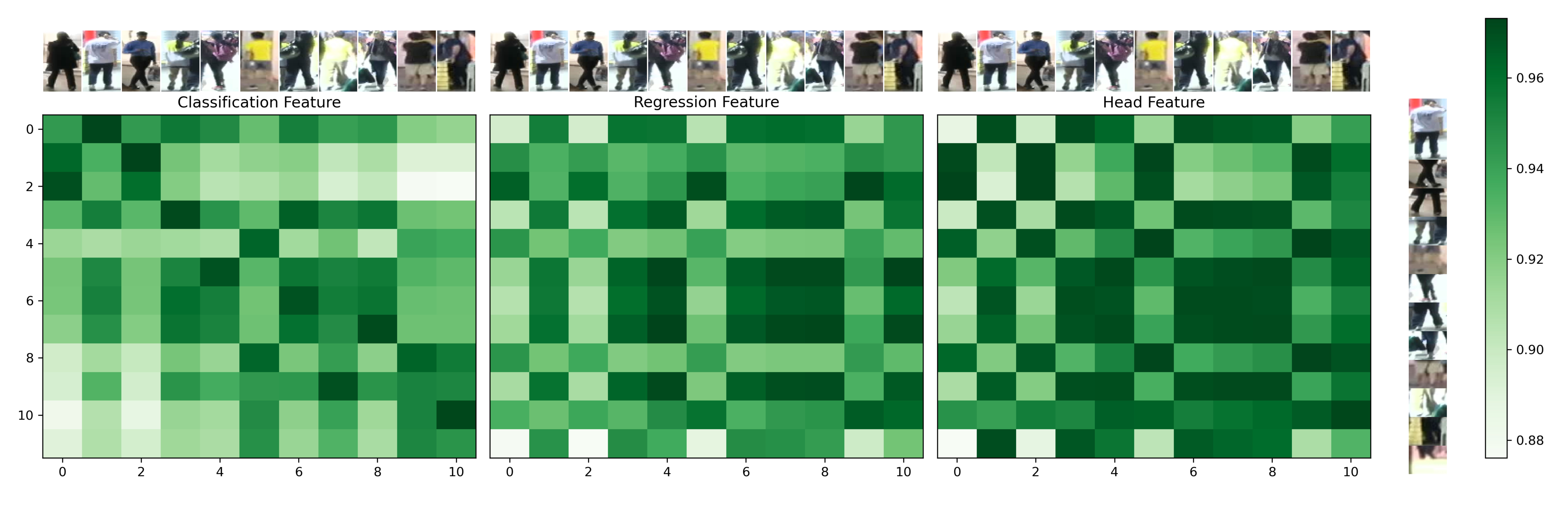}
	\caption{llustration of different detector features for {Re-ID} task.
    }
	\label{fig:heat_figure}
\end{figure}

We also investigate the effectiveness of different detector features for {Re-ID} task. 
Figure~\ref{fig:heat_figure} presents a comparison of features at different scales extracted respectively from the classification branch, regression branch, and head keypoints of the detector. 
The multi-scale features extracted from neck module are first processed through stem convolutions in the head module. 
These features are then passed through task-specific convolution branches, producing feature maps of dimensions $1\times320\times100\times100$ for P3, $1\times320\times50\times50$ for P4, and $1\times320\times25\times25$ for P5. 
These features are extracted according to the bounding boxes and transformed into a unified feature map of size $320\times7\times7$, which is further flattened into a $15680$ dimensional feature vector for subsequent processing.

To evaluate the discriminative power of these features, we select two adjacent frames from the dataset and extract target features from each frame. The similarity between these features is computed using cosine distance and visualized through similarity matrices.
The figure displays similarity matrices between targets across two frames from the MOT17-10 sequence (3-th and 4-th frames), where darker colors indicate higher similarity. From these results, we can draw the following observations:
i) Classification features outperform the others in terms of discriminative ability and are effective at distinguishing different targets.
ii) Head keypoint features show relatively weaker performance, similar to that of regression features. This may be due to their optimization objective focusing on localization, which makes it difficult to capture useful appearance information.
iii) Although regression features perform similarly to head keypoint features, they appear to be slightly more discriminative. This may be attributed to the nature of their optimization objective, which potentially retains more informative cues.
The suboptimal performance of both regression and head keypoint features might also be related to the size of the extracted feature maps, which is set to $7\times7 $ in this study. This aspect warrants further investigation.

\begin{figure} [htbp]
	\centering
	\includegraphics[width=8.0 cm]{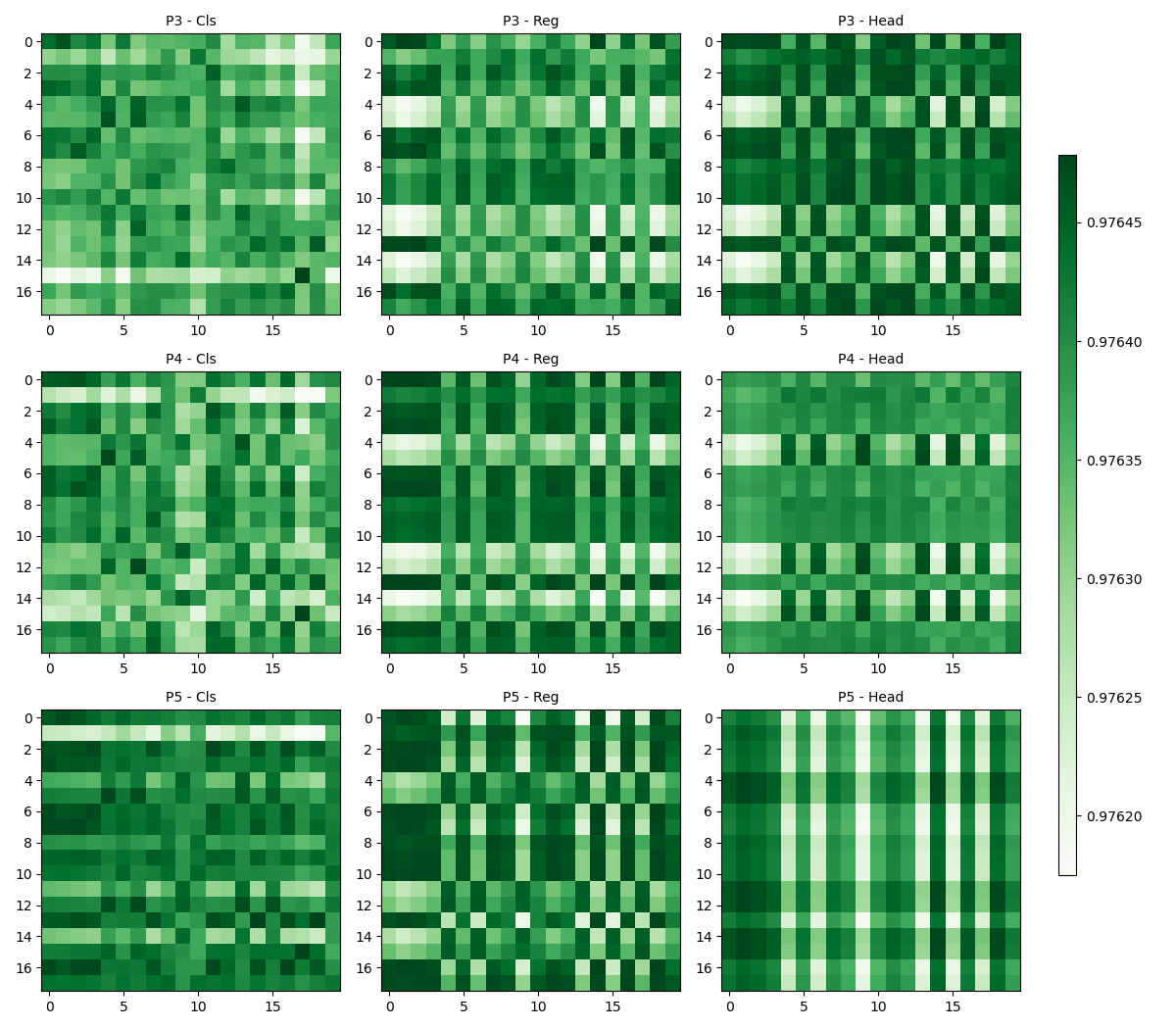}
	\caption{Illustration of different scale features for {Re-ID} task.
    }
	\label{fig:heat_feats}
\end{figure}

To better investigate the impact of feature scale on similarity discrimination, we conducted additional comparative experiments by sampling features from the FT03 dataset. The comparison results are illustrated in Figure~\ref{fig:heat_feats}.

As shown in the figure, features from P3 exhibit superior performance in similarity comparison tasks compared to those from P4 and P5, suggesting that lower-level features may retain more discriminative spatial details beneficial for appearance matching. Furthermore, comparisons among classification features, regression features from full-body bounding boxes, and head region features further support our previous findings: classification features serve as a strong representation for appearance similarity, while regression and head features are more sensitive to spatial proximity.

Interestingly, the head features behave similarly to the regression features extracted from the body, but yield a clearer decision boundary. This aligns with our understanding that head positions are generally more stable and less affected by posture variations, thereby offering advantages in scenarios involving occlusion or densely populated targets.

\begin{figure} [htbp]
	\centering
	\includegraphics[width=8.0 cm]{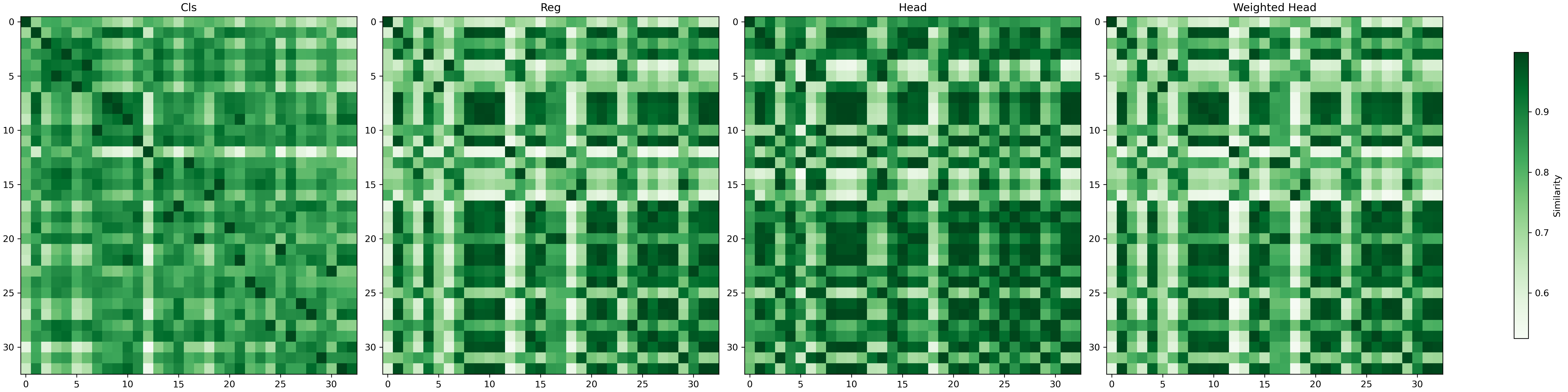}
	\caption{Illustration of different P3 features for {Re-ID} task.
    The weighted head features enhance the original box-based features through Gaussian weighting centered on the head keypoint position. 
    This weighting mechanism emphasizes the spatial information around the head region, resulting in more focused and discriminative head-specific feature representations.
    }
	\label{fig:heat_fig_weight}
\end{figure}

Based on the experimental results, we observe that shallow-layer (P3) classification features are more suitable for target assignment tasks. 
However, other feature representations also demonstrate distinct advantages in specific scenarios, particularly the head features. 
To better leverage the effectiveness of head keypoint, we further process them by applying Gaussian weighting based on head keypoint coordinates to obtain enhanced detector feature representations as shown in Figure~\ref{fig:heat_fig_weight}.

\subsection{Effect of Detector Features in Tracking Tasks}

To evaluate the effectiveness of our detector's feature representation, we conducted comparative studies focusing solely on appearance-based tracking. 
This tested tracker is a simplified SORT framework (denoted as sSORT), which utilizes appearance distance for cost matrix computation.
Table~\ref{tab:only_feat} presents the tracking results on MOT17-04 when using only appearance cost, isolating the impact of feature quality from other tracking components.

\begin{table}[htbp]
\centering
\setlength{\tabcolsep}{3pt}
\caption{Comparative Study on Appearance-based Tracking}
\label{tab:only_feat}
\begin{tabular}{l|ccccc}
\hline
Method & MOTA $\uparrow$ & IDF1 $\uparrow$ & FP $\downarrow$ & FN $\downarrow$ & IDs $\downarrow$ \\
\hline
 sSORT+ $F_\text{motion}$ & 71.91 & 77.85 & 127 & 13180 & 51 \\
\hline
 + $F_\text{cls}$  & 70.33 & 66.40 & 506 & 13477 & 127 \\
 + $F_\text{reg}$  & 67.32 & 53.22 & 1002 & 14015 & 525 \\
 + $F_\text{head}$ & 59.56 & 37.64 & 2337 & 15430 & 1465 \\
 + $F_\text{cls}$ + $F_\text{reg}$ & \textbf{70.58} & 65.75 & \textbf{458} & \textbf{13416} & \textbf{117} \\
 + $F_\text{cls}$ + $F_\text{g\_reg}$ & 70.34 & \textbf{66.41} & 505 & 13476 & 126 
\\
\hline
\end{tabular}
\end{table}

These quantitative results align well with our qualitative observations from the visualization analysis. The strong tracking performance achieved using only classification feature $F_\text{cls}$ and regression feature $F_\text{reg}$, without relying on dedicated appearance features, demonstrates the detector's inherent capability in providing discriminative representations. 
This finding suggests promising potential for detectors to serve as effective feature extractors for {Re-ID} tasks.
Furthermore, when considering head features alone, the discriminative capability of target representation is notably limited, which aligns with our theoretical expectations. This reduced discriminative power can be attributed to the fact that head feature optimization primarily focuses on positional information while containing relatively sparse appearance features.

Furthermore, these results demonstrate potential opportunities for improvement through the integration of classification features, regression features, and head keypoint information. Compared to using classification features ($F_\text{cls}$) alone, combining multiple feature types $F_\text{cls}+F_\text{reg}$ or $F_\text{g\_reg}$ ($F_\text{g\_reg}$ indicates Gaussian weighted $F_\text{reg}$ with head postion) shows room for enhanced performance. Future work could explore feature aggregation strategies that first decouple detector branches before merging them to obtain more tracking-oriented feature representations. This approach may better capture the complementary strengths of different feature types while maintaining their distinct characteristics.

\subsection{Analysis of Trajectory Complementary Methods}

\begin{figure} [htbp]
	\centering
	\includegraphics[width=8.0 cm]{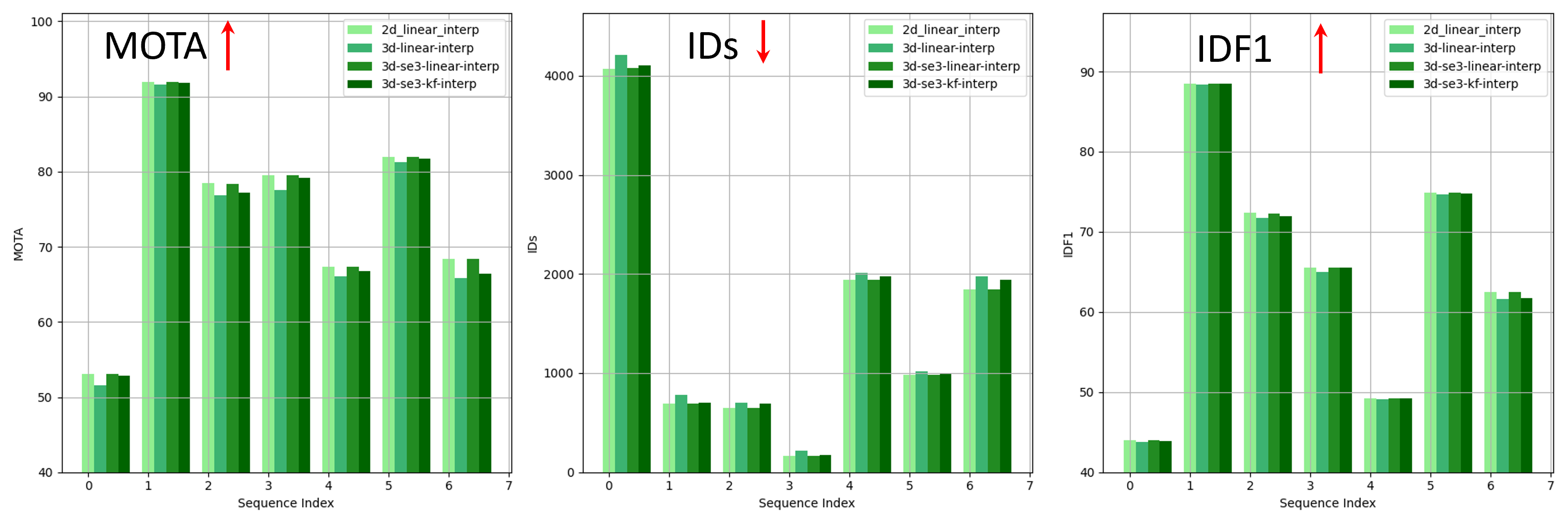}
	\caption{Performances of different post-process with interpolation.
    The figure illustrates four distinct interpolation approaches: 2D linear interpolation, 3D linear interpolation, 3D SE(3) linear interpolation, and 3D SE(3) Kalman filter-based interpolation.
    }
	\label{fig:intep}
\end{figure}

To evaluate the effectiveness of trajectory completion methods, we conducted comparative experiments using four different approaches: 2D linear interpolation, 3D linear interpolation, SE(3)-based 3D interpolation, and 3D Kalman filter-based interpolation. The experimental results are illustrated in the Figure~\ref{fig:intep}.

We compared the performance of these four interpolation methods across three metrics. Overall, these methods exhibited consistent performance on these major tracking metrics. 2D linear interpolation performed the best, followed by Kalman filter-based 3D SE(3) interpolation and linear SE(3) interpolation, with linear 3D interpolation ranking last.

Generally, 2D interpolation delivered better results. This is understandable because our implementation used pseudo-depth estimation to simplify the 2D-to-3D lifting process. This led to inaccurate representations in 3D space, which explains why 3D interpolation methods underperformed compared to their 2D counterparts.

Among the 3D interpolation methods, the performance aligned with expectations. SE(3)-based interpolation outperformed methods without such modeling, and the Kalman filter-based approach surpassed linear interpolation. These improvements introduced more realistic constraints for trajectory completion.

\subsection{Benchmark Evaluation}

We conduct comprehensive benchmark evaluations on three challenging datasets: MOT17, MOT20, and FT25. 
The proposed FocusTrack is compared against several state-of-the-art tracking methods, including SORT~\cite{bewley2016simple}, OCSORT~\cite{cao2023observation}, Bot-SORT~\cite{aharon2022bot}, and ByteTrack~\cite{zhang2022bytetrack}. 
To ensure fair comparison across different tracking methods, all trackers were evaluated using identical detection results. 
For MOT17 and MOT20 datasets, detections were generated using the detector proposed in ByteTrack. 
For the FT25 dataset, ground truth labels provided by the simulator were used as detection inputs to eliminate detection errors and focus purely on tracking performance.
Besides, all trackers in this evaluation were tested without incorporating {Re-ID} features, focusing solely on their core tracking capabilities.

On the MOT17 dataset, FocusTrack achieves competitive performance comparable to Bot-SORT, with a MOTA of 78.3\% and IDF1 of 71.0\%. The method demonstrates strong tracking capabilities with 15,410 false positives and 1,248 false negatives, while maintaining robust identity preservation with only 1248 identity switches.
FocusTrack demonstrates second-best performance on most sequences of MOT17.
Similar performance patterns are observed on the MOT20 dataset, where FocusTrack matches Sort and ByteTrack's results across all metrics. This consistency across different datasets validates the robustness and reliability of our approach under varying tracking conditions.
The evaluation on FT25 further corroborates these findings, with FocusTrack maintaining its strong performance aligned with ByteTrack. This consistent performance across multiple benchmarks demonstrates the generalization capability of our method.

These comprehensive results establish FocusTrack as a robust tracking solution that effectively minimizes detection errors while maintaining high identity preservation accuracy across diverse tracking scenarios. 
The consistent performance across different datasets validates the effectiveness of our proposed head-enhanced representation and association strategy.

\begin{table}[htbp]
    \centering
    \setlength{\tabcolsep}{2pt}
    \setlength{\itemsep}{1pt}
    \caption{Benchmark evaluation on MOT17 dataset}
    \label{tab:benchmark_mot17}
    \begin{tabular}{c|l|ccccc}
    \hline
    SEQ & Method & MOTA↑ & IDF1↑ & FP↓ & FN↓ & IDs↓ \\
    \hline
    \multirow{6}{*}{02} & SORT~\cite{bewley2016simple} & \textbf{54.8} & 44.7 & \textbf{3165} & \textbf{4873} & \underline{361} \\
    & OCSORT~\cite{cao2023observation} & 53.3 & \textbf{46.3} & 3284 & 5037 & 363 \\
    & ByteTrack~\cite{zhang2022bytetrack} & 54.4 & 43.6 & 3208 & 4902 & 366 \\
    & Bot-SORT~\cite{aharon2022bot} & {54.4} & \underline{45.7} & 3211 & 4901 & \textbf{355} \\
    & HybridSORT~\cite{yang2024hybrid} & 53.6 & 41.8 & 3220 & 5028 & 379 \\
    & FocusTrack~(Ours) & \underline{54.6} & 44.3 & \underline{3179} & \underline{4887} & {375} \\ \hline
    \multirow{6}{*}{04} & SORT~\cite{bewley2016simple} & \textbf{91.1} & \textbf{88.4} & \underline{557} & 3600 & \textbf{72} \\
    & OCSORT~\cite{cao2023observation} & {90.9} & 84.8 & 589 & 3644 & 98 \\
    & ByteTrack~\cite{zhang2022bytetrack} & \textbf{91.1} & \underline{88.1} & {561} & \underline{3603} & {77} \\
    & Bot-SORT~\cite{aharon2022bot} & \textbf{91.1} & 87.6 & \textbf{555} & \textbf{3594} & \textbf{72} \\
    & HybridSORT~\cite{yang2024hybrid} & {90.9} & 86.1 & 577 & 3654 & 98 \\
    & FocusTrack~(Ours) & \underline{91.0} & \underline{88.1} & 569 & 3615 & \underline{73} \\ \hline
    \multirow{6}{*}{05} & SORT~\cite{bewley2016simple} & {77.4} & {72.0} & \underline{505} & \underline{942} & \underline{96} \\
    & OCSORT~\cite{cao2023observation} & 73.3 & 70.0 & 574 & 1056 & 217 \\
    & ByteTrack~\cite{zhang2022bytetrack} & {77.4} & \underline{72.2} & 525 & \underline{942} & \underline{96} \\
    & Bot-SORT~\cite{aharon2022bot} & \textbf{78.0} & \textbf{75.8} & 519 & \textbf{931} & \textbf{72} \\
    & HybridSORT~\cite{yang2024hybrid} & 74.6 & 66.1 & 546 & 1036 & 173 \\
    & FocusTrack~(Ours) & \underline{77.7} & 71.9 & \textbf{497} & 944 & 102 \\ \hline
    \multirow{6}{*}{09} & SORT~\cite{bewley2016simple} & \underline{77.1} & \underline{65.0} & \underline{111} & \textbf{1059} & {48} \\
    & OCSORT~\cite{cao2023observation} & 76.0 & 57.5 & 120 & 1078 & 80 \\
    & ByteTrack~\cite{zhang2022bytetrack} & 76.7 & 64.9 & 121 & 1069 & 50 \\
    & Bot-SORT~\cite{aharon2022bot} & 77.0 & \textbf{66.4} & 120 & \underline{1064} & \underline{42} \\
    & HybridSORT~\cite{yang2024hybrid} & 76.1 & 62.2 & 122 & 1090 & 61 \\
    & FocusTrack~(Ours) & \textbf{77.3} & 62.6 & \textbf{103} & \underline{1064} & \textbf{41} \\ \hline
    \multirow{6}{*}{10} & SORT~\cite{bewley2016simple} & {69.1} & 50.2 & \underline{1357} & \underline{2206} & \underline{346} \\
    & OCSORT~\cite{cao2023observation} & 65.4 & 44.0 & 1465 & 2448 & 532 \\
    & ByteTrack~\cite{zhang2022bytetrack} & 68.6 & {51.8} & 1403 & 2254 & 371 \\
    & Bot-SORT~\cite{aharon2022bot} & \textbf{71.2} & \textbf{63.0} & \underline{1357} & \textbf{2126} & \textbf{216} \\
    & HybridSORT~\cite{yang2024hybrid} & 67.9 & 52.6 & \textbf{1345} & 2360 & 411 \\
    & FocusTrack~(Ours) & \underline{69.5} & \underline{52.8} & 1369 & \underline{2206} & \underline{346} \\ \hline
    \multirow{6}{*}{11} & SORT~\cite{bewley2016simple} & 82.7 & 75.3 & \underline{751} & \underline{741} & \underline{66} \\
    & OCSORT~\cite{cao2023observation} & 82.0 & \underline{78.4} & 786 & 818 & 99 \\
    & ByteTrack~\cite{zhang2022bytetrack} & 82.7 & 72.2 & 788 & 769 & 78 \\
    & Bot-SORT~\cite{aharon2022bot} & \textbf{83.8} & \textbf{84.7} & 766 & \textbf{716} & \textbf{50} \\
    & HybridSORT~\cite{yang2024hybrid} & {83.1} & 77.1 & \textbf{736} & 782 & 78 \\
    & FocusTrack~(Ours) & \underline{83.5} & 76.0 & \underline{751} & \underline{741} & \underline{66} \\ \hline
    \multirow{6}{*}{13} & SORT~\cite{bewley2016simple} & {70.0} & \underline{62.9} & \textbf{1255} & \underline{1952} & {265} \\
    & OCSORT~\cite{cao2023observation} & 58.7 & 47.9 & 1429 & 2466 & 918 \\
    & ByteTrack~\cite{zhang2022bytetrack} & {70.0} & 62.3 & \underline{1270} & \underline{1952} & {265} \\
    & Bot-SORT~\cite{aharon2022bot} & \textbf{71.5} & \textbf{68.5} & 1282 & \textbf{1854} & \textbf{183} \\
    & HybridSORT~\cite{yang2024hybrid} & 64.9 & 53.2 & 1284 & 2177 & 622 \\
    & FocusTrack~(Ours) & \underline{70.3} & 61.9 & \textbf{1255} & 1953 & \underline{245} \\ \hline
    \multirow{6}{*}{Total} & SORT~\cite{bewley2016simple} & {78.2} & \underline{71.0} & \underline{7752} & {15453} & {1254} \\
    & OCSORT~\cite{cao2023observation} & 75.9 & 67.3 & 8247 & 16547 & 2307 \\
    & ByteTrack~\cite{zhang2022bytetrack} & 78.0 & 70.6 & 7876 & 15491 & 1303 \\
    & Bot-SORT~\cite{aharon2022bot} & \textbf{78.6} & \textbf{74.0} & 7810 & \textbf{15186} & \textbf{990} \\
    & HybridSORT~\cite{yang2024hybrid} & 77.0 & 68.6 & 7830 & 16127 & 1822 \\
    & FocusTrack~(Ours) & \underline{78.3} & \underline{71.0} & \textbf{7723} & \underline{15410} & \underline{1248} \\
    \hline 
    \end{tabular}
    \end{table}
    
\begin{table}[htbp]
\centering
\setlength{\tabcolsep}{1pt}
\caption{Benchmark evaluation on MOT20 dataset}
\label{tab:benchmark_mot20}
\begin{tabular}{c|l|ccccc}
\hline
SEQ & Method & MOTA↑ & IDF1↑ & FP↓ & FN↓ & IDs↓ \\
\hline
\multirow{5}{*}{01} & SORT~\cite{bewley2016simple} & \textbf{76.3} & \underline{68.6} & \underline{900} & \underline{3644} & \textbf{157} \\
& OCSORT~\cite{cao2023observation} & 74.8 & 61.1 & 953 & 3773 & 281 \\
& ByteTrack~\cite{zhang2022bytetrack} & \underline{76.2} & 66.0 & 905 & 3654 & \underline{165} \\
& HybridSORT~\cite{yang2024hybrid} & 74.5 & 56.3 & 952 & 3792 & 321 \\
& FocusTrack~(Ours) & \underline{76.2} & \textbf{68.8} & \textbf{894} & \textbf{3642} & 180 \\
\hline
\multirow{5}{*}{02} & SORT~\cite{bewley2016simple} & \textbf{77.5} & \underline{58.7} & \textbf{3341} & \textbf{30385} & \textbf{1161} \\
& OCSORT~\cite{cao2023observation} & 76.6 & 53.2 & 3652 & 30974 & 1593 \\
& ByteTrack~\cite{zhang2022bytetrack} & \underline{77.4} & \textbf{59.0} & \underline{3361} & \underline{30409} & \underline{1177} \\
& HybridSORT~\cite{yang2024hybrid} & 76.5 & 47.5 & 3475 & 31005 & 1912 \\
& FocusTrack~(Ours) & 77.3 & 58.2 & \underline{3361} & 30463 & 1266 \\
\hline
\multirow{5}{*}{03} & SORT~\cite{bewley2016simple} & \textbf{88.6} & \textbf{85.2} & \textbf{8599} & \textbf{26626} & \textbf{684} \\
& OCSORT~\cite{cao2023observation} & 88.1 & 79.7 & 8811 & 27195 & 1169 \\
& ByteTrack~\cite{zhang2022bytetrack} & {88.2} & {84.8} & 9188 & 27174 & {751} \\
& HybridSORT~\cite{yang2024hybrid} & 88.0 & 65.1 & \underline{8604} & 27313 & 1686 \\
& FocusTrack~(Ours) & \underline{88.5} & \underline{85.1} & 8637 & \underline{26663} & \underline{744} \\
\hline
\multirow{5}{*}{05} & SORT~\cite{bewley2016simple} & \textbf{81.3} & \textbf{63.8} & \textbf{22158} & \textbf{95622} & \textbf{3151} \\
& OCSORT~\cite{cao2023observation} & 80.8 & 60.4 & 22595 & 97201 & 4377 \\
& ByteTrack~\cite{zhang2022bytetrack} & {80.9} & {63.0} & 23270 & 96638 & \underline{3243} \\
& HybridSORT~\cite{yang2024hybrid} & 80.4 & 44.7 & \underline{22181} & 98057 & 6726 \\
& FocusTrack~(Ours) & \underline{81.2} & \underline{63.3} & 22231 & \underline{95723} & 3382 \\
\hline
\multirow{5}{*}{Total} & SORT~\cite{bewley2016simple} & \textbf{82.7} & \textbf{69.3} & {34998} & \textbf{156277} & \textbf{5153} \\
& OCSORT~\cite{cao2023observation} & 82.1 & 64.9 & 36011 & 159143 & 7420 \\
& ByteTrack~\cite{zhang2022bytetrack} & {82.4} & {68.7} & 36724 & 157875 & \underline{5336} \\
& HybridSORT~\cite{yang2024hybrid} & 81.8 & 51.1 & \underline{35212} & 160167 & 10645 \\
& FocusTrack~(Ours) & \underline{82.6} & \underline{68.9} & \textbf{35123} & \underline{156501} & 5572 \\
\hline
\end{tabular}
\end{table}

\begin{table}[htbp]
\centering
\setlength{\tabcolsep}{2pt}
\caption{Benchmark evaluation on FT25 dataset.}
\label{tab:benchmark_ft25}
\begin{tabular}{c|l|ccccc}
\hline
SEQ & Method & MOTA↑ & IDF1↑ & FP↓ & FN↓ & IDs↓ \\
\hline
\multirow{5}{*}{03} & SORT~\cite{bewley2016simple} & 63.6 & {78.8} & 949 & \underline{5027} & \underline{7} \\
& OCSORT~\cite{cao2023observation} & 63.7 & 76.1 & \underline{911} & 5028 & 29 \\
& ByteTrack~\cite{zhang2022bytetrack} & \textbf{68.0} & \textbf{80.7} & \textbf{134} & 5121 & \textbf{6} \\
& HybridSORT~\cite{yang2024hybrid} & 61.3 & 76.0 & 1120 & 5171 & 62 \\
& FocusTrack~(Ours) & \underline{63.9} & \underline{78.9} & 931 & \textbf{4992} & 16 \\
\hline
\multirow{5}{*}{04} & SORT~\cite{bewley2016simple} & \underline{89.4} & \textbf{87.6} & {15} & \underline{1117} & \textbf{19} \\
& OCSORT~\cite{cao2023observation} & 89.2 & \underline{87.2} & \textbf{4} & 1136 & 29 \\
& ByteTrack~\cite{zhang2022bytetrack} & 88.2 & 87.1 & 79 & 1181 & \underline{20} \\
& HybridSORT~\cite{yang2024hybrid} & 88.4 & 84.4 & 61 & 1165 & 32 \\
& FocusTrack~(Ours) & \textbf{89.5} & \textbf{87.6} & \underline{6} & \textbf{1110} & 21 \\
\hline
\multirow{5}{*}{Total} & SORT~\cite{bewley2016simple} & {73.8} & {82.4} & 964 & \underline{6144} & \textbf{26} \\
& OCSORT~\cite{cao2023observation} & {73.8} & 80.7 & \underline{915} & 6164 & 58 \\
& ByteTrack~\cite{zhang2022bytetrack} & \textbf{76.0} & \textbf{83.4} & \textbf{213} & 6302 & \textbf{26} \\
& HybridSORT~\cite{yang2024hybrid} & 72.1 & 79.5 & 1181 & 6336 & 94 \\
& FocusTrack~(Ours) & \underline{74.0} & \underline{82.6} & 937 & \textbf{6102} & \underline{37} \\
\hline
\end{tabular}
\end{table}

\section{Potential Limitations}

This work proposes an effective representation and association strategy for pedestrian tracking under occlusion, demonstrating promising performance in challenging scenarios. 
However, the proposed head-enhanced representation method may encounter limitations under the following specific conditions:
\begin{itemize}
    \item Horizontal viewpoints: When the camera observes pedestrians from a nearly horizontal or side view, the visibility of the head region is reduced, leading to less reliable representations.
    \item Close-range targets: When pedestrians are too close to the camera, the head region may fall outside the field of view or exhibit significant scale variation, making accurate localization difficult.
    \item Occlusion by other scene elements: The presence of occluding objects such as vehicles, baggage, or other pedestrians can degrade the quality and discriminative power of the head-based features.
\end{itemize}

In such scenarios, short-term associations may be insufficient to maintain consistent trajectories. Therefore, more reliable long-term association strategies, such as {MHT}, are required to enhance robustness in the presence of severe occlusions or temporary target disappearance.

\bibliographystyle{elsarticle-num}
%
\bibliography{cas-refs.bib}

\end{document}